\newif\ifdraft\draftfalse

\newif\ifinlineref\inlinereffalse
\inlinereftrue

\newif\ifrevmarkerA\revmarkerAfalse

\documentclass[oneside,a4paper]{article}

\usepackage[sort]{natbib}
\usepackage{fullpage}

\usepackage[T1]{fontenc}
\usepackage[utf8]{inputenc}
\usepackage{amsmath}
\usepackage{amsthm}
\usepackage{amsfonts}
\usepackage{setspace}
\usepackage{graphicx}
\usepackage{url}
\usepackage{amssymb}
\usepackage{multirow}
\usepackage{paralist}
\usepackage{listings}
\usepackage{tabularx}
\usepackage{rotating}
\usepackage{booktabs}
\usepackage{tikz}
\usetikzlibrary{shapes,fit,arrows,calc,positioning}
\usepackage{subcaption}

\ifdraft
  \usepackage{todonotes}
\else
  \usepackage[disable]{todonotes}
\fi

\ifrevmarkerA
  \newcommand{\revA}[1]{{\color{blue}#1}}
  \newcommand{\revAc}[0]{\color{blue}}
\else
  \newcommand{\revA}[1]{{#1}}
  \newcommand{\revAc}[0]{}
\fi

\newtheorem{example}{Example}

\newcommand{\naf}{\ensuremath{\mathbf{not}\,}}

\newcommand{\lifs}{\ensuremath{\,{\leftarrow}\,}}
\newcommand\quo[1]{`#1'}
\newcommand{\mi}[1]{\ensuremath{\mathit{#1}}}

\newcommand{\ins}{\,{\in}\,}

\newcommand{\gts}{\,{>}\,}

\newcommand{\eqs}{\,{=}\,}

\newcommand{\sig}{%
\begin{tikzpicture}[overlay]
\node[xshift=5pt,yshift=8pt] {|\raisebox{-2pt}{*}};
\end{tikzpicture}}

\newcommand{\HeadlinesResults}[1]{
\begin{sidewaystable}
\small
\singlespacing%
\centering%
\revA{%
\begin{tabular}{ccccccccccc}
\toprule
 &  &       & \multicolumn{4}{c@{\quad}}{\bf S1} &\multicolumn{4}{c@{\quad}}{\bf S2}\\
\cmidrule(lr){4-7} \cmidrule(lr){8-11}
\bf Size & 		\boldmath$Pr$   &\multicolumn{1}{c}{\boldmath$So$} & \bf CV  & \multicolumn{3}{c@{\quad}}{\bf T} & \bf CV &\multicolumn{3}{c@{\quad}}{\bf T}\\
\cmidrule(lr{2cm}){4-5} \cmidrule(lr){5-7} \cmidrule(lr{2cm}){8-9} \cmidrule(lr){9-11}
&				   & 				   & \bf F1  & \bf P & \bf R & \bf F1            & \bf F1 & \bf P & \bf R & \bf F1 \\
\midrule
100 & 0 & 172.8$\pm$46.2 & 66.3$\pm$10.1 & 63.0$\pm$2.2 & 64.9$\pm$3.3 & 59.4$\pm$3.3 & 70.7$\pm$14.2 & 65.5$\pm$2.4 & 64.6$\pm$2.7 & 60.5$\pm$2.6 \\

    & 1 & 10.9$\pm$5.0 & 71.1$\pm$13.3\sig & 69.4$\pm$2.0 & 67.1$\pm$2.0 & 63.8$\pm$2.2\sig & 69.3$\pm$15.7 & 67.3$\pm$0.5 & 66.2$\pm$1.4 & 62.4$\pm$1.0\sig \\

    & 2 & 0.3$\pm$0.8 & 73.1$\pm$8.0 & 69.3$\pm$0.7 & 69.2$\pm$1.1 & 65.0$\pm$0.4\sig & 66.5$\pm$15.4 & 65.9$\pm$1.5 & 68.2$\pm$0.5 & 62.7$\pm$1.1 \\

    & 3 & 0.0$\pm$0.0 & 65.9$\pm$5.9 & 66.6$\pm$3.4 & 69.7$\pm$1.7 & 63.0$\pm$2.9 & 65.1$\pm$15.6 & 64.7$\pm$0.9 & 68.5$\pm$0.3 & 61.6$\pm$0.5 \\
\midrule
500 & 0 & 31954.4$\pm$7057.7 & 39.4$\pm$18.1 & 50.9$\pm$9.8 & 34.8$\pm$18.7 & 35.7$\pm$14.0 & 39.2$\pm$12.7 & 53.2$\pm$8.0 & 38.4$\pm$14.1 & 38.9$\pm$11.7 \\

    & 1 & 17855.1$\pm$6866.0 & 39.1$\pm$14.9 & 51.9$\pm$9.2 & 39.0$\pm$17.9 & 38.3$\pm$13.9 & 40.7$\pm$12.6 & 53.4$\pm$8.7 & 40.0$\pm$14.7 & 39.7$\pm$11.9 \\

    & 2 & 6238.5$\pm$1530.8 & 55.8$\pm$10.5\sig & 59.6$\pm$4.2 & 57.0$\pm$9.2 & \,53.2$\pm$6.8\,\sig & 53.0$\pm$14.3\sig & 59.4$\pm$5.7 & 52.0$\pm$11.9 & 49.4$\pm$9.5 \\

    & 3 & 4260.9$\pm$792.4 & 52.5$\pm$11.4 & 59.2$\pm$5.0 & 52.4$\pm$11.8 & \,49.6$\pm$9.3\,\sig & 59.2$\pm$7.8 & 62.1$\pm$2.9 & 58.5$\pm$4.6 & 54.7$\pm$3.5 \\

    & 4 & 1598.6$\pm$367.1 & \,65.7$\pm$3.8\,\sig & 65.2$\pm$3.3 & 66.3$\pm$3.0 & \,61.1$\pm$3.0\,\sig & 67.1$\pm$8.4\sig & 63.3$\pm$2.0 & 64.7$\pm$4.1 & 59.7$\pm$3.0\sig \\

    & 5 & 1117.2$\pm$211.3 & 67.0$\pm$4.6 & 66.8$\pm$3.1 & 67.8$\pm$3.1 & 62.9$\pm$3.0 & 73.3$\pm$7.3\sig & 65.5$\pm$2.3 & 66.4$\pm$3.6 & 62.1$\pm$2.7\sig \\

    & 6 & 732.6$\pm$130.4 & 69.7$\pm$4.0 & 67.5$\pm$1.9 & 69.5$\pm$2.4 & 64.3$\pm$2.1 & 71.7$\pm$5.7 & 65.3$\pm$1.6 & 67.4$\pm$4.4 & 62.7$\pm$2.9\sig \\

    & 7 & 561.4$\pm$81.8 & 68.2$\pm$4.5 & 67.2$\pm$1.9 & 70.5$\pm$1.5 & 64.5$\pm$1.8 & 71.2$\pm$7.1 & 66.5$\pm$1.0 & 68.0$\pm$1.7 & 63.6$\pm$1.0 \\

    & 8 & 475.0$\pm$142.3 & 68.9$\pm$4.5 & 67.0$\pm$2.6 & 69.0$\pm$5.8 & 63.8$\pm$4.4 & 71.8$\pm$5.7 & 67.2$\pm$1.3 & 68.2$\pm$2.4 & 64.0$\pm$1.8 \\

    & 9 & 312.7$\pm$111.2 & 69.3$\pm$6.2 & 68.1$\pm$2.5 & 70.6$\pm$2.5 & 65.4$\pm$2.6 & 71.2$\pm$5.5 & 66.6$\pm$1.4 & 67.5$\pm$2.3 & 63.3$\pm$2.0 \\

    & 10 & 220.3$\pm$59.9 & 67.8$\pm$4.5 & 67.3$\pm$2.1 & 70.9$\pm$2.8 & 65.0$\pm$2.4 & 73.4$\pm$6.7 & 66.1$\pm$1.7 & 68.6$\pm$2.1 & 63.7$\pm$1.6 \\
\bottomrule
\end{tabular}
}%
\caption{Experimental Results for Headlines Dataset\revA{, where * indicates statistical significance ($p < 0.05$).
Additionally, for Size $=500$, the F1 scores for all pruning values $Pr > 1$ are significantly better than $Pr = 0$ ($p < 0.05$).}}
\label{table:Headlines}
\end{sidewaystable}
}

\newcommand{\ImagesResults}[1]{
\begin{sidewaystable}
\small
\singlespacing%
\centering%
\revA{%
\begin{tabular}{ccccccccccc}
\toprule
 &  &       & \multicolumn{4}{c@{\quad}}{\bf S1} &\multicolumn{4}{c@{\quad}}{\bf S2}\\
\cmidrule(lr){4-7} \cmidrule(lr){8-11}
\bf Size & 		\boldmath$Pr$   &\multicolumn{1}{c}{\boldmath$So$} & \bf CV  & \multicolumn{3}{c@{\quad}}{\bf T} & \bf CV &\multicolumn{3}{c@{\quad}}{\bf T}\\
\cmidrule(lr{2cm}){4-5} \cmidrule(lr){5-7} \cmidrule(lr{2cm}){8-9} \cmidrule(lr){9-11}
&				   & 				   & \bf F1  & \bf P & \bf R & \bf F1            & \bf F1 & \bf P & \bf R & \bf F1 \\
\midrule
100 & 0 & 0.5$\pm$1.0 & 81.8$\pm$12.7 & 66.4$\pm$15.5 & 74.3$\pm$0.7 & 73.7$\pm$0.7 & 73.9$\pm$0.7 & 60.1$\pm$9.5 & 60.1$\pm$9.6 & 60.1$\pm$9.5 \\

    & 1 & 0.0$\pm$0.0 & 80.9$\pm$14.4 & 64.5$\pm$10.8 & 72.7$\pm$1.4 & 72.1$\pm$1.4 & 72.3$\pm$1.4 & 50.2$\pm$5.6 & 50.0$\pm$5.6 & 50.1$\pm$5.6 \\

    & 2 & 0.0$\pm$0.0 & 78.2$\pm$15.3 & 64.5$\pm$13.7 & 69.2$\pm$1.4 & 68.9$\pm$0.8 & 68.9$\pm$1.2 & 47.5$\pm$1.8 & 47.3$\pm$1.8 & 47.4$\pm$1.8 \\

    & 3 & 0.0$\pm$0.0 & 72.7$\pm$14.2 & 66.4$\pm$16.7 & 67.0$\pm$0.5 & 67.8$\pm$0.5 & 67.1$\pm$0.5 & 47.3$\pm$1.5 & 47.1$\pm$1.5 & 47.2$\pm$1.5 \\
\midrule
500 & 0 & 3797.3$\pm$1019.9 & 47.6$\pm$8.6 & 45.9$\pm$12.5 & 47.1$\pm$8.8 & 46.2$\pm$8.9 & 46.4$\pm$8.9 & 45.0$\pm$12.9 & 45.5$\pm$12.8 & 44.6$\pm$13.2 \\

     & 1 & 670.1$\pm$153.1 & 64.2$\pm$8.2\sig & 68.1$\pm$7.4 & 57.1$\pm$11.1 & 56.1$\pm$11.1\sig & 56.4$\pm$11.1\sig & 63.1$\pm$9.2 & 63.1$\pm$9.5 & ~62.9$\pm$9.4~\sig \\

     & 2 & 286.2$\pm$90.2 & 69.5$\pm$4.9\sig & 73.8$\pm$7.1 & 66.4$\pm$6.6 & \:65.6$\pm$6.6\:\sig & \:65.8$\pm$6.6\:\sig & 68.4$\pm$6.0 & 68.4$\pm$6.0 & ~68.2$\pm$6.0~\sig \\

     & 3 & 159.1$\pm$36.4 & 70.9$\pm$6.8 & 70.1$\pm$7.0 & 66.0$\pm$7.6 & 65.4$\pm$7.8 & 65.4$\pm$7.7 & 69.8$\pm$3.7 & 69.7$\pm$3.6 & 69.6$\pm$3.7 \\

     & 4 & 83.4$\pm$25.8 & 74.7$\pm$5.7 & 68.8$\pm$6.4 & 70.2$\pm$2.0 & 69.6$\pm$1.9 & 69.7$\pm$1.9 & 67.0$\pm$7.2 & 66.7$\pm$7.2 & 66.7$\pm$7.2 \\

     & 5 & 23.8$\pm$11.0 & 74.2$\pm$6.6 & 70.7$\pm$4.7 & 71.9$\pm$1.5 & 71.1$\pm$1.4\sig & 71.3$\pm$1.4 & 71.0$\pm$1.7 & 70.9$\pm$1.8 & 70.8$\pm$1.8 \\

     & 6 & 10.8$\pm$4.5 & 75.3$\pm$5.9 & 73.2$\pm$4.5 & 71.7$\pm$0.4 & 71.0$\pm$0.4 & 71.2$\pm$0.4 & 71.1$\pm$0.8 & 71.1$\pm$0.8 & 70.9$\pm$0.8 \\

     & 7 & 3.4$\pm$3.6 & 74.4$\pm$5.9 & 72.1$\pm$4.2 & 71.2$\pm$0.3 & 70.5$\pm$0.3 & 70.7$\pm$0.3 & 69.7$\pm$1.4 & 69.7$\pm$1.4 & 69.6$\pm$1.4 \\

     & 8 & 1.5$\pm$1.4 & 74.5$\pm$5.3 & 72.3$\pm$5.8 & 71.2$\pm$0.0 & 70.4$\pm$0.0 & 70.6$\pm$0.0 & 68.6$\pm$0.8 & 68.6$\pm$0.7 & 68.4$\pm$0.7 \\

     & 9 & 1.2$\pm$1.4 & 74.5$\pm$5.3 & 71.9$\pm$5.8 & 71.2$\pm$0.0 & 70.4$\pm$0.0 & 70.6$\pm$0.0 & 68.4$\pm$0.5 & 68.3$\pm$0.5 & 68.2$\pm$0.5 \\

     & 10 & 0.7$\pm$0.8 & 74.2$\pm$5.2 & 71.8$\pm$5.5 & 70.9$\pm$0.9 & 70.1$\pm$0.9 & 70.4$\pm$0.9 & 68.6$\pm$0.0 & 68.5$\pm$0.0 & 68.3$\pm$0.0 \\
\bottomrule
\end{tabular}
}%
\caption{Experimental Results for Images Dataset\revA{, where * indicates statistical significance ($p < 0.05$).
Additionally, for Size $=500$, the F1 scores for all pruning values $Pr > 0$ are significantly better than $Pr = 0$ ($p < 0.05$).}}
\label{table:Images}
\end{sidewaystable}
}

\newcommand{\StudentAnsResults}[1]{
\begin{sidewaystable}
\small
\singlespacing%
\centering%
\revA{%
\begin{tabular}{cccccccccc}
\toprule
 &  &       & \bf S1+S2 & \multicolumn{3}{c@{\quad}}{\bf S1} &\multicolumn{3}{c@{\quad}}{\bf S2}\\
\cmidrule(lr{2cm}){4-5} \cmidrule(lr){5-7} \cmidrule(lr){8-10}
\bf Size & 		\boldmath$Pr$   &\multicolumn{1}{c}{\boldmath$So$} & \bf CV  & \multicolumn{3}{c@{\quad}}{\bf T} & \multicolumn{3}{c@{\quad}}{\bf T}\\
\cmidrule(lr{2cm}){4-5} \cmidrule(lr){5-7} \cmidrule(lr){8-10}
&				   & 				   & \bf F1  & \bf P & \bf R & \bf F1            & \bf P & \bf R & \bf F1 \\
\midrule
100 & 0 & 93.2$\pm$22.6 & 66.1$\pm$12.9 & 69.3$\pm$1.5 & 63.2$\pm$3.2 & 61.0$\pm$2.6 & 89.3$\pm$3.0 & 80.1$\pm$0.7 & 80.3$\pm$1.7 \\

    & 1 & 5.3$\pm$4.6   & 67.0$\pm$11.5 & 67.9$\pm$1.3 & 61.6$\pm$1.8 & 59.4$\pm$1.7 & 87.7$\pm$1.0 & 79.7$\pm$0.8 & 79.5$\pm$1.0 \\

    & 2 & 0.0$\pm$0.0   & 65.0$\pm$10.8 & 67.7$\pm$0.7 & 64.9$\pm$1.4 & 61.2$\pm$0.8 & 86.3$\pm$1.5 & 80.2$\pm$0.5 & 78.4$\pm$1.4 \\

    & 3 & 0.0$\pm$0.0   & 64.8$\pm$10.4 & 66.8$\pm$0.4 & 63.0$\pm$1.4 & 59.9$\pm$0.5 & 86.6$\pm$1.5 & 80.7$\pm$0.3 & 78.9$\pm$1.4 \\
\midrule
500 & 0 & 20723.5$\pm$3996.9 & 36.3$\pm$10.6 & 54.2$\pm$5.5 & 39.8$\pm$13.2 & 38.7$\pm$9.9 & 51.2$\pm$10.8 & 37.2$\pm$15.0 & 36.0$\pm$12.4 \\

    & 1 & 6643.4$\pm$1131.1 & \,49.3$\pm$8.7\,\sig & 60.0$\pm$4.9 & 51.3$\pm$10.1 & 48.7$\pm$7.2 & 62.9$\pm$9.9 & 53.4$\pm$15.6 & 52.1$\pm$13.8\sig \\

    & 2 & 4422.2$\pm$734.7 & 54.5$\pm$9.7 & 62.6$\pm$2.4 & 60.6$\pm$8.1 & 56.1$\pm$5.5\sig & 66.4$\pm$7.0 & 59.9$\pm$11.7 & 57.6$\pm$10.8\sig \\

    & 3 & 2782.2$\pm$626.9 & 57.5$\pm$10.6 & 62.2$\pm$3.0 & 62.4$\pm$8.7 & 56.7$\pm$5.7 & 67.6$\pm$10.8 & 62.2$\pm$15.5 & 59.4$\pm$14.5 \\

    & 4 & 1541.6$\pm$311.3 & \,65.5$\pm$4.1\,\sig & 66.8$\pm$2.8 & 70.5$\pm$2.5 & 63.5$\pm$2.4\sig & 78.9$\pm$2.0 & 79.5$\pm$2.0 & \,76.0$\pm$2.0\,\sig \\

    & 5 & 1072.4$\pm$155.5 & 63.6$\pm$7.8 & 66.1$\pm$2.7 & 67.6$\pm$5.1 & 61.7$\pm$3.5 & 80.8$\pm$3.4 & 77.1$\pm$3.3 & 74.5$\pm$3.0 \\

    & 6 & 789.1$\pm$158.0 & 64.8$\pm$6.5 & 65.5$\pm$1.9 & 64.2$\pm$4.9 & 59.4$\pm$3.6 & 83.3$\pm$2.7 & 75.9$\pm$4.0 & 74.4$\pm$3.3 \\

    & 7 & 634.7$\pm$184.0 & 66.3$\pm$7.8 & 65.9$\pm$2.2 & 64.6$\pm$3.5 & 59.7$\pm$2.6 & 82.9$\pm$3.4 & 75.2$\pm$5.2 & 74.2$\pm$4.0 \\

    & 8 & 449.8$\pm$87.4 & 63.9$\pm$6.6 & 65.0$\pm$2.5 & 64.5$\pm$6.5 & 59.1$\pm$4.5 & 80.3$\pm$4.4 & 74.2$\pm$8.2 & 72.2$\pm$6.8 \\

    & 9 & 317.0$\pm$89.7 & 63.9$\pm$6.4 & 64.1$\pm$2.4 & 64.3$\pm$4.0 & 58.3$\pm$3.4 & 82.3$\pm$4.3 & 74.9$\pm$5.5 & 72.9$\pm$5.3 \\

    & 10 & 225.5$\pm$45.7 & 63.4$\pm$5.1 & 66.6$\pm$1.7 & 65.3$\pm$2.6 & 60.1$\pm$1.8 & 82.4$\pm$5.2 & 74.1$\pm$8.0 & 72.7$\pm$7.4 \\
\bottomrule
\end{tabular}
}%
\caption{Experimental Results for Answers-Students Dataset\revA{, where * indicates statistical significance ($p < 0.05$).
Additionally, for Size $=500$, the F1 scores for all pruning values $Pr > 0$ are significantly better than $Pr = 0$ ($p < 0.05$).}}
\label{table:StudentAns}
\end{sidewaystable}
}

\newcommand{\FinalResults}[1]{
\begin{table}
\small
\singlespacing%
\centering%
\setlength{\tabcolsep}{0.4em}
\begin{tabularx}{0.98\textwidth}{cccc@{\quad}l@{\qquad}lll@{~}c@{\qquad}lll@{~}c}
\toprule
\multicolumn{4}{c}{\multirow{2}{*}{\bf Data}} &
\multirow{2}{*}{\bf System} &
\multicolumn{3}{c@{}}{\bf S1} & &
\multicolumn{3}{c@{}}{\bf S2} & \\
\cmidrule(l{-2mm}r{5mm}){6-9}
\cmidrule(l{-1mm}r{2mm}){10-13}
\multicolumn{4}{c}{} & {}
& \bf P & \bf R & \bf F1 & \bf Rank
& \bf P & \bf R & \bf F1 & \bf Rank \\

\midrule
\multicolumn{4}{c}{\multirow{7}{*}{\rotatebox{90}{Headlines}}} & Baseline
& 60.5 & 36.6 & 37.6 &
& 63.6 & 42.5 & 42.8 & \\

\multicolumn{4}{c}{} & DTSim
& 72.5 & 74.3 & 71.3 &*
& 72.1 & 74.3 & 70.5 &* \\

\multicolumn{4}{c}{} & FBK-HLT-NLP
& 63.6 & 51.3 & 51.5 &
& 57.1 & 51.1 & 48.3 & \\

\multicolumn{4}{c}{} & IISCNLP - Run1
& 61.9 & 68.5 & 61.4 &
& 61.1 & 65.7 & 60.1 & \\

\multicolumn{4}{c}{} & IISCNLP - Run2
& 67.6 & 68.5 & 64.5 &***
& 71.4 & 71.9 & 68.9 &** \\

\multicolumn{4}{c}{} & Inspire - Manual
& 64.5 & 70.4 & 62.4 &
& 64.3 & 68.4 & 62.2 & \\

\multicolumn{4}{c}{} & Inspire - Learned
& \revAc 68.1$\pm$2.5 & \revAc 70.6$\pm$2.5 & \revAc 65.4$\pm$2.6 &\revAc **
& \revAc 67.2$\pm$1.3 & \revAc 68.2$\pm$2.4 & \revAc 64.0$\pm$1.8 &\revAc ***\quad \\
\midrule

\multicolumn{4}{c}{\multirow{7}{*}{\rotatebox{90}{Images}}} & Baseline
& 19.0 & 15.7 & 16.4 &
& 13.6 & 17.5 & 13.5 & \\

\multicolumn{4}{c}{} & DTSim
& 77.8 & 77.4 & 77.5 &*
& 79.5 & 79.1 & 79.2 &* \\

\multicolumn{4}{c}{} & FBK-HLT-NLP
& 41.0 & 39.2 & 38.8 &
& 40.5 & 43.1 & 40.8 & \\

\multicolumn{4}{c}{} & IISCNLP - Run1
& 61.6 & 60.9 & 60.7 &
& 66.1 & 66.2 & 65.9 & \\

\multicolumn{4}{c}{} & IISCNLP - Run2
& 65.8 & 65.6 & 65.4 &
& 67.7 & 67.2 & 67.3 & \\

\multicolumn{4}{c}{} & Inspire - Manual
& 74.5 & 74.2 & 74.2 &**
& 73.8 & 73.6 & 73.6 &** \\

\multicolumn{4}{c}{} & Inspire - Learned
& \revAc 66.4$\pm$15.5 & \revAc 74.3$\pm$0.7 & \revAc 73.7$\pm$0.7 &\revAc ***
& \revAc 71.1$\pm$0.8  & \revAc 71.1$\pm$0.8 & \revAc 70.9$\pm$0.8 &\revAc *** \\
\midrule

\multicolumn{4}{c}{\multirow{7}{*}{\rotatebox{90}{Answers-Students}}} & Baseline
& 62.1 & 30.9 & 34.6 &
& 59.2 & 33.4 & 36.6 & \\

\multicolumn{4}{c}{} & DTSim
& 78.5 & 73.6 & 72.5 &*
& 83.3 & 79.2 & 77.8 &** \\

\multicolumn{4}{c}{} & FBK-HLT-NLP
& 70.3 & 52.5 & 52.8 &
& 72.4 & 59.1 & 59.3 & \\

\multicolumn{4}{c}{} & IISCNLP - Run1
& 67.9 & 63.9 & 60.7 & ***
& 65.7 & 55.0 & 54.0 & \\

\multicolumn{4}{c}{} & IISCNLP - Run2
& 63.0 & 59.8 & 56.9 &
& 66.2 & 52.5 & 52.8 & \\

\multicolumn{4}{c}{} & Inspire - Manual
& 66.8 & 64.4 & 59.7 &
& 71.2 & 62.5 & 62.1 &*** \\

\multicolumn{4}{c}{} & Inspire - Learned
&\revAc  66.8$\pm$2.8 &\revAc  70.5$\pm$2.5 &\revAc  63.5$\pm$2.4 &\revAc **
&\revAc  89.3$\pm$3.0 &\revAc  80.1$\pm$0.7 &\revAc  80.3$\pm$1.7 &\revAc * \\
\bottomrule

\end{tabularx}
\caption{Comparison with systems from SemEval 2016 Task 2. The number of stars shows the rank of the system.}
\label{table:final}
\end{table}
}

\newcommand{\DataSplit}[1]{
\begin{table}
  \small
  \singlespacing%
  \centering%
  \revA{%
  \begin{tabular*}{\textwidth}{@{\extracolsep{\fill}}@{\quad}cr@{\quad}lc@{\quad}c@{\quad}}
  \toprule
  \bf Dataset & \multicolumn{2}{l}{\bf Cross-Validation Set} & \multicolumn{2}{c}{\bf Test Set} \\
		& Size & Examples 				& \bf S1  & \bf S2 \\
  \midrule
  \multirow{4}{*}{H/I} & 100 & S1 first 110 & all & * \\

		       & 500 & S1 first 550 & all & * \\

		       & 100 & S2 first 110 & *      & all \\
		       & 500 & S2 first 550 & *      & all \\
  \midrule
  \multirow{2}{*}{A-S} & 100 & S1 first 55\hphantom{0} + S2 first 55\hphantom{0} & all & all \\
		       & 500 & S1 first 275 + S2 first 275 & all & all \\
  \bottomrule
  \end{tabular*}
  }%
  \caption{%
  Dataset partitioning \revA{for 11-fold cross-validation experiments.
  Size indicates the \emph{training set size} in cross-validation.}
  Fields marked with * are not applicable,
  because we do not evaluate hypotheses learned from
  the S1 portion of the
  Headlines (H) and Images (I) datasets on the (independent)
  S2 portion of these datasets and vice versa.
  For the Answers-Students (A-S) dataset we need to merge S1 and S2
  to obtain a training set size of 500 from the (small) SemEval Training dataset.}
  \label{table:Split}
\end{table}
}

\newcommand{\xfw}[1]{%
\tikzstyle{bblock} = [rectangle, draw, line width=0.45mm, text width=7em, text centered, rounded corners, minimum height=3em,font=\small]
\tikzstyle{block} = [rectangle, draw,text width=7em, text centered, rounded corners, minimum height=3em,font=\small]
\tikzstyle{dblock} = [rectangle, draw, line width=0.45mm, text width=7em, text centered, rounded corners, minimum height=3em,font=\small, dotted]
\tikzstyle{line} = [draw, -latex', line width=0.45mm, black]
\tikzstyle{l} = [draw, -latex']
\tikzstyle{line node} = [draw, fill=white, font=\small]
\tikzstyle{c} = [rectangle, draw, inner sep=0.25cm, dashed]
\begin{figure}
\small
\centering
\scalebox{.8}
{
\begin{tikzpicture}[
  node distance = 1cm,
  every node/.style={rectangle,fill=white}
  ]
    \node [block]                 	  (A)  {Abduction};
    \node [left=1cm of A]     (Ex) {\begin{tabular}{@{}r@{}}
      Examples $E$ \\
      Background Knowledge $B$ \\
      Mode Bias $M$ (Head)
      \end{tabular}};
    \node [block, below=of A]        (D)  {Deduction};
    \node [left=1cm of D]     (M)  {Mode Bias $M$ (Body)};
    \node [dblock, below=of D]        (G)  {Generalisation};
    \node [block, below=of G]        (I)  {Induction};
    \node [below=0.3cm of I]        (H)  {Hypothesis};
    \node [bblock, right=3.0cm of D,yshift=-1cm]  (Gn) {Generalisation (counting)};
    \node [bblock, below=of Gn]       (P)  {Pruning};

    \path [l] (Ex) -- (A);
    \path [l] (A) -- node {$\Delta$ (Kernet Set)} (D);
    \path [l] (D) -- node {ground K program} (G);
    \path [l] (G) -- node {non-ground K' program} (I);
    \path [l] (I) -- (H);
    \path [l] (M) -- (D);

    \path [line] (D) -- ++(4,0) node {ground K program} -| (Gn);

    \path [line] (Gn) -- node {non-ground K' program with support counts} (P);
    \path [line] (P) -- ++(0,-1) node[anchor=east,xshift=-0.5cm] {subset of K'} |- (I);

\node[left=1cm of G,yshift=-0.75cm] {%
\begin{tabular}{cc}
 \raisebox{2pt}{\tikz{\draw[line width=0.45mm, black,dotted] (0,0) -- (5mm,0);}}&Replaced\\
 \raisebox{2pt}{\tikz{\draw[line width=0.45mm, black] (0,0) -- (5mm,0);}}&Modified
\end{tabular}};
   \end{tikzpicture}%
}
   \caption{XHAIL architecture. The dotted line shows the replaced module with our version represented by
   the thick solid line.}%
    \label{figXhail}%
\end{figure}%
}

\newcommand{\work}[1]{
\tikzstyle{b} = [rectangle, draw, node distance=1.1cm, text width=6em, text centered, rounded corners, minimum height=4em, thick]
\tikzstyle{c} = [rectangle, draw, inner sep=0.25cm, dashed]
\tikzstyle{l} = [draw, -latex',thick]

\begin{figure}
  \small
  \centering
\begin{tikzpicture}[auto]
    \node [b] (process) {Stanford Core-NLP tools};
    \node [c, fit=(process), inner sep=0.25cm] (container1) {};
    \node [b, right=of container1] (xhail) {ILP tool XHAIL};
    \node [c, fit=(xhail), inner sep=0.25cm] (container2) {};
    \node [b, right=of container2] (chunk) {Chunking with ASP};
    \node [c, fit=(chunk), inner sep=0.25cm] (container3) {};

    \path [l] (container1) -- (container2);
    \path [l] (container2) -- (container3);

    \node at (container1.north west) [above right] {Preprocessing};
    \node at (container2.north west) [above right] {Learning};
    \node at (container3.north west) [above right] {Testing};

    \end{tikzpicture}
  \caption{General overview of our framework}
  \label{table:Work}
\end{figure}
}

\newcommand{\rules}[1]{
\begin{table*}
\footnotesize%
\singlespacing%
\centering%
{%
\begin{tabular}{lccc}
\toprule
\textbf{Rules} & \textbf{H} & \textbf{I} & \textbf{A-S} \\
\midrule
split(V) :- token(V), pos(c\_VBD,V).& X & X & X \\
split(V) :- token(V), nextpos(c\_IN,V).& X & X & X \\
split(V) :- token(V), nextpos(c\_VBZ,V).& X & X & X \\
split(V) :- token(V), pos(c\_VB,V).& X &  & X \\
split(V) :- token(V), nextpos(c\_TO,V).& X &  & X \\
split(V) :- token(V), nextpos(c\_VBD,V).& X &  & X \\
split(V) :- token(V), nextpos(c\_VBP,V).&  & X & X \\
split(V) :- token(V), pos(c\_VBZ,V), nextpos(c\_DT,V).&  & X & X \\
split(V) :- token(V), pos(c\_NN,V), nextpos(c\_RB,V).&  & X & X \\
split(V) :- token(V), pos(c\_NNS,V).& X &  &  \\
split(V) :- token(V), pos(c\_VBP,V).& X &  &  \\
split(V) :- token(V), pos(c\_VBZ,V).& X &  &  \\
split(V) :- token(V), pos(c\_c,V).& X &  &  \\
split(V) :- token(V), nextpos(c\_POS,V).& X &  &  \\
split(V) :- token(V), nextpos(c\_VBN,V).& X &  &  \\
split(V) :- token(V), nextpos(c\_c,V).& X &  &  \\
split(V) :- token(V), pos(c\_PRP,V).&  & X &  \\
split(V) :- token(V), pos(c\_RP,V).&  & X &  \\
split(V) :- token(V), pos(c\_p,V).&  & X &  \\
split(V) :- token(V), nextpos(c\_p,V).&  & X &  \\
split(V) :- token(V), pos(c\_CC,V), nextpos(c\_VBG,V).&  & X &  \\
split(V) :- token(V), pos(c\_NN,V), nextpos(c\_VBD,V).&  & X &  \\
split(V) :- token(V), pos(c\_NN,V), nextpos(c\_VBG,V).&  & X &  \\
split(V) :- token(V), pos(c\_NN,V), nextpos(c\_VBN,V).&  & X &  \\
split(V) :- token(V), pos(c\_NNS,V), nextpos(c\_VBG,V).&  & X &  \\
split(V) :- token(V), pos(c\_RB,V), nextpos(c\_IN,V).&  & X &  \\
split(V) :- token(V), pos(c\_VBG,V), nextpos(c\_DT,V).&  & X &  \\
split(V) :- token(V), pos(c\_VBG,V), nextpos(c\_JJ,V).&  & X &  \\
split(V) :- token(V), pos(c\_VBG,V), nextpos(c\_PRPd,V).&  & X &  \\
split(V) :- token(V), pos(c\_VBG,V), nextpos(c\_RB,V).&  & X &  \\
split(V) :- token(V), pos(c\_VBZ,V), nextpos(c\_IN,V).&  & X &  \\
split(V) :- token(V), pos(c\_EX,V).&  &  & X  \\
split(V) :- token(V), pos(c\_RB,V).&  &  & X  \\
split(V) :- token(V), pos(c\_VBG,V).&  &  & X  \\
split(V) :- token(V), pos(c\_WDT,V).&  &  & X  \\
split(V) :- token(V), pos(c\_WRB,V).&  &  & X  \\
split(V) :- token(V), nextpos(c\_EX,V).&  &  & X  \\
split(V) :- token(V), nextpos(c\_MD,V). &  &  & X  \\
split(V) :- token(V), nextpos(c\_VBG,V).&  &  & X  \\
split(V) :- token(V), nextpos(c\_RB,V).&  &  & X  \\
split(V) :- token(V), pos(c\_IN,V), nextpos(c\_NNP,V).&  &  & X  \\
split(V) :- token(V), pos(c\_NN,V), nextpos(c\_WDT,V).&  &  & X  \\
split(V) :- token(V), pos(c\_NN,V), nextpos(c\_IN,V).&  &  & X  \\
split(V) :- token(V), pos(c\_NNS,V), nextpos(c\_IN,V).&  &  & X  \\
split(V) :- token(V), pos(c\_NNS,V), nextpos(c\_VBP,V).&  &  & X  \\
split(V) :- token(V), pos(c\_RB,V), nextpos(c\_DT,V).&  &  & X  \\
\bottomrule
\end{tabular}
}
\caption{%
Rules in the best hypotheses obtained from training on 500 sentences (S1),
where X marks the presence of the rule in a given dataset.
}
\label{table:Rules}
\end{table*}
}

\title{Improving Scalability of Inductive Logic Programming\\
via Pruning and Best-Effort Optimisation}

\author{Mishal Kazmi$^1$, Peter Sch\"{u}ller$^{2,C}$, and Y\"{u}cel Sayg{\i}n$^1$\\[1ex]
$^1$ Faculty of Engineering and Natural Science,\\
Sabanci University,\\
Istanbul, Turkey \\
{\tt \{mishalkazmi,ysaygin\}@sabanciuniv.edu} \\
$^2$ Faculty of Engineering,\\
Marmara University,\\
Istanbul, Turkey \\
{\tt peter.schuller@marmara.edu.tr} / {\tt schueller.p@gmail.com}}

\date{\small%
~\\[1ex]
Technical Report: manuscript accepted for publication
at \emph{Expert Systems With Applications} (Elsevier). \\
\copyright\ 2017.
This manuscript version is made available under the CC-BY-NC-ND 4.0 license.
{\tt http://creativecommons.org/licenses/by-nc-nd/4.0/}\enspace.}

\begin{document}

\maketitle

\begin{abstract}
Inductive Logic Programming (ILP) combines rule-based and
statistical artificial intelligence methods, by learning a
hypothesis comprising a set of rules given background knowledge
and constraints for the search space.  We focus on extending the
XHAIL algorithm for ILP which is based on Answer Set Programming
and we evaluate our extensions using the Natural Language
Processing application of sentence chunking.  With respect
to processing natural language, ILP can cater for the constant
change in how we use language on a daily basis.  At the same
time, ILP does not require huge amounts of training examples such
as other statistical methods and produces interpretable results,
that means a set of rules, which can be analysed and tweaked if
necessary.  As contributions we extend XHAIL with (i) a pruning
mechanism within the hypothesis generalisation algorithm which
enables learning from larger datasets, (ii) a better usage of
modern solver technology using recently developed optimisation
methods, and (iii) a time budget that permits the usage of
suboptimal results.  We evaluate these improvements on the task
of sentence chunking using three datasets from a recent SemEval
competition.  Results show that our improvements allow for
learning on bigger datasets with results that are of similar
quality to state-of-the-art systems on the same task.  Moreover,
we compare the hypotheses obtained on datasets to gain insights
on the structure of each dataset.

\end{abstract}

\section{Introduction}
Inductive Logic Programming (ILP) \citep{muggleton1994inductive}
is a formalism where
a set of logical rules is learned from a set of examples
and a background knowledge theory.
By combining rule-based and statistical artificial intelligence,
ILP overcomes the brittleness of pure logic-based approaches
and the lack of interpretability of models of most statistical methods such as neural networks or support vector machines.
We here focus on ILP that is based on Answer Set Programming (ASP) as our underlying logic programming language
because we aim to apply ILP to
Natural Language Processing (NLP) applications
such as Machine Translation, Summarization,
Coreference Resolution, or Parsing
that require nonmonotonic reasoning with exceptions
and complex background theories.

\revA{%
In our work, we apply ILP to the NLP task of sentence chunking.
Chunking, also known as `shallow parsing', is the identification of short phrases such as noun phrases
which mainly rely on Part of Speech (POS) tags.
}%
In our experiments on
sentence chunking \citep{tjong2000introduction}
we encountered several problems with state-of-the-art
ASP-based ILP systems
XHAIL \citep{ray2009nonmonotonic},
ILED \citep{katzouris2015incremental},
and ILASP2 \citep{law2015learning}.
XHAIL and ILASP2 showed scalability issues already
with 100 sentences as training data.
ILED is designed to be highly scalable
but failed in the presence of simple inconsistencies
in examples.
We decided to investigate the issue in the XHAIL system,
which is open-source and documented well,
and we made the following observations:
\begin{enumerate}[(i)]
\item XHAIL only terminates if it finds a provably optimal hypothesis,
\item the hypothesis search is done over all potentially beneficial rules that are supported by at least one example, and
\item XHAIL contains redundancies in hypothesis search and uses outdated ASP technology.
\end{enumerate}

In larger datasets, observation (i) is unrealistic,
because finding a near-optimal solution is much easier
than proving optimality of the best solution,
moreover in classical machine learning suboptimal solutions
obtained via non-exact methods routinely provide
state-of-the-art results.
Similarly, observation (ii) makes it harder to find a hypothesis,
and it generates an overfitting hypotheses
which contains rules that are only required for a single example.
Observation (iii) points out an engineering problem
that can be remedied with little theoretical effort.

To overcome the above issues,
we modified the XHAIL algorithm and software,
and we performed experiments on a simple NLP chunking task
to evaluate our modifications.
\smallskip

In detail, we make the following contributions.
\begin{itemize}
\item
  We extend XHAIL with best-effort optimisation
  using the newest ASP optimisation technology
  of unsat-core optimisation \citep{Andres2012}
  with stratification \citep{Ansotegui2013maxsat,Alviano2015aspopt}
  and core shrinking \citep{Alviano2016coreshrinking}
  using the WASP2 \citep{DBLP:conf/lpnmr/AlvianoDFLR13,DBLP:conf/lpnmr/AlvianoDLR15}
  solver
  and the Gringo \citep{Gebser2011gringo3}
  grounder.
  We also extend XHAIL to provide information about the optimality
  of the hypothesis.
\item
  We extend the XHAIL algorithm with a parameter $\mi{Pr}$
  for pruning,
  such that XHAIL searches for hypotheses
  without considering rules that are supported by
  fewer than $\mi{Pr}$ examples.
\item
  We eliminate several redundancies in XHAIL
  by changing its internal data structures.
\item
  We describe a framework for chunking with ILP,
  based on preprocessing with Stanford Core NLP \citep{manning2014stanford} tools.
\item
  We experimentally analyse the relationship between the
  pruning parameter, number of training examples,
  and prediction score on the sentence chunking \citep{tjong2000introduction} subtask
  of iSTS at SemEval 2016 \citep{companionpaper}.
\item
  We discuss the best hypothesis found for each
  of the three datasets in the SemEval task,
  and we discuss what can be learned
  about the dataset from these hypotheses.
\end{itemize}
Only if we use all the above modifications together,
XHAIL becomes applicable in this chunking task.
By learning a hypothesis from 500 examples,
we can achieve results competitive with
state-of-the-art systems used in the SemEval 2016 competition.

Our extensions and modifications of the XHAIL software
are available in a public fork of the official
XHAIL Git repository \citep{xhailfork}.

In Section \ref{prelim} we provide an overview of logic programming and ILP.
Section \ref{pastwork} gives an account of related work
and available ILP tools.
In Section \ref{secExtendingXHAIL}
we describe the XHAIL system and our extensions of pruning,
best-effort optimisation, and further improvements.
Section \ref{chunk} gives details of our representation of
the chunking task.
In Section \ref{results} we discuss empirical experiments
and results.
We conclude in Section \ref{conc} with a brief outlook on future work.

\section{Background} \label{prelim}

We next introduce logic programming and based on that inductive logic programming.

\subsection{Logic Programming}
\label{secLP}
A logic programs theory normally comprises of an alphabet (variable, constant, quantifier, etc), vocabulary, logical symbols, a set of axioms and inference rules
\citep{lloyd2012foundations}.
A logic programming system consists of two portions: the logic and control.
Logic describes what kind of problem needs to be solved and control is how that
problem can be solved. An ideal of logic programming is for it to be purely declarative.
The popular Prolog \citep{clocksin2003programming} system
evaluates rules using resolution, which makes the result of a Prolog program
depending on the order of its rules and on the order of the bodies of its rules.
Answer Set Programming (ASP) \citep{brewka2011answer,Gebser2012aspbook}
is a more recent logic programming formalism,
featuring more declarativity than Prolog
by defining semantics based on Herbrand models \citep{Gelfond1988}.
Hence the order of rules and the order
of the body of the rules does not matter in ASP.
Most ASP programs follow the Generate-Define-Test structure \citep{Lifschitz2002}
to
(i) generate a space of potential solutions,
(ii) define auxiliary concepts, and
(iii) test to invalidate solutions using constraints or incurring a cost on non-preferred solutions.

An ASP program consists of rules of the following structure:
\begin{align*}
a \lifs b_1,\ \ldots, b_m, \naf b_{m+1},\ldots,\naf b_n
\end{align*}
where $a$, $b_i$ are atoms from a first-order language, $a$ is the head and $b_1,\ldots,\naf b_n$ is the body of the rule, and \naf\ is negation as failure.
Variables start with capital letters, facts (rules without body condition) are written as \quo{$a.$} instead of  \quo{$a \lifs $}.
Intuitively $a$ is true if all positive body atoms are true and no negative body atom is true.

The formalism can be understood more clearly by considering the following sentence as a simple example:
\begin{align*}
\textit{Computers are normally fast machines unless they are old.}
\end{align*}
This would be represented as a logical rule as follows:
\begin{align*}
\textit{fastmachine(X)} \lifs \textit{computer(X)}, \textit{\naf old(X).}
\end{align*}
where $X$ is a variable, $\mi{fastmachine}$, $\mi{computer}$, and $\mi{old}$ are predicates,
and $\mi{old(X)}$ is a negated atom.

Adding more knowledge results in a change of a previous understanding, this is common in human reasoning.
Classical First Order Logic does not allow such non-monotonic reasoning,
however, ASP was designed as a commonsense reasoning formalism:
a program has zero or more answer sets as solutions,
adding knowledge to the program can remove answer sets as well as produce new ones.
Note that ASP semantics rule out self-founded truths in answer sets.
We use the ASP formalism due to its flexibility and declarativity.
For formal details and a complete description of syntax and semantics see the ASP-Core-2 standard \citep{Calimeri2012}.
ASP has been applied to several problems related to
Natural Language Processing, see for example
\citep{Schwitter2012,Schuller2013aspccg,Schuller2014winograd,Sharma2015,Schuller2016aspfoa,mitra2016addressing}.
An overview of applications of ASP in general can be found
in \citep{Erdem2016aimag}.

\subsection{Inductive Logic Programming}

Processing natural language based on hand-crafted rules is impractical
because human language is constantly evolving,
partially due to the human creativity of language use.
An example of this was recently noticed on UK highways where they advised drivers, \quo{Don't Pok\'emon Go and drive}. Pok\'emon Go is being informally
used here as a verb even though it was only introduced as a game a few weeks before the sign was put up.
To produce robust systems,
it is necessary to use statistical models of language.
These models are often pure Machine Learning (ML) estimators
without any rule components \citep{manning1999foundations}.
ML methods work very well in practice,
however, they usually do not provide a way for explaining why
a certain prediction was made,
because
they represent the learned knowledge in big matrices of real numbers.
Some popular classifiers used for processing natural language
include Naive Bayes, Decision Trees, Neural Networks,
and Support Vector Machines (SVMs) \citep{dumais1998inductive}.

In this work, we focus on an approach that combines rule-based methods
and statistics and provides interpretable learned models:
\emph{Inductive Logic Programming} (ILP).
ILP is differentiated from ML techniques by its use
of an expressive representation language and its ability
to make use of logically encoded background knowledge
\citep{muggleton1994inductive}.
An important advantage of ILP over ML techniques such as neural networks is,
that a hypothesis can be made readable by translating it
into piece of English text.
Furthermore,
if annotated corpora of sufficient size are not available or too expensive to produce,
deep learning or other data intense techniques are not applicable.
However, we can still learn successfully with ILP.
\smallskip

Formally, ILP takes as input a set of examples $E$,
a set $B$ of background knowledge rules, and
a set of mode declarations $M$, also called mode bias.
As output, ILP aims to produce a set of rules $H$ called hypothesis
which entails $E$ with respect to $B$.
The search for $H$ with respect to $E$ and $B$ is restricted by $M$,
which defines a language that limits the shape of rules in the hypothesis candidates
and therefore the complexity of potential hypotheses.

\begin{example}
\label{exMini}
Consider the following example ILP instance $(M,E,B)$
\citep{ray2009nonmonotonic}.
\begin{align}
&M = \left \{
  \begin{tabular}{@{}l@{}}
  \#modeh flies(+bird).\\
  \#modeb penguin(+bird).\\
  \#modeb \naf penguin(+bird).
  \end{tabular}
\right \}
\label{eqM} \\
&E = \left \{
  \begin{tabular}{@{}l@{}}
\#example flies(a). \\
\#example flies(b). \\
\#example flies(c). \\
\#example \naf flies(d).
  \end{tabular}
\right \}
\label{eqE} \\
&B = \left \{
  \begin{tabular}{@{}l@{}}
  bird(X) :- penguin(X). \\
  bird(a). \\
  bird(b). \\
  bird(c). \\
  penguin(d).
  \end{tabular}
\right \} \\
\intertext{%
Based on this, an ILP system would ideally find the following hypothesis.
}%
  &H = \left \{
    \begin{tabular}{@{\,}l@{\,}}
    flies(X) :- bird(X), \naf penguin(X).
    \end{tabular}
  \right \}
  \label{eqExHypothesis}
\end{align}
\end{example}

\section{Related Work} \label{pastwork}

Inductive Logic Programming (ILP) is a rather multidisciplinary field which extends to domains such as computer science, artificial intelligence, and
bioinformatics. Research done in ILP has been greatly impacted by Machine Learning (ML), Artificial Intelligence (AI) and relational
databases. Quite a few surveys \citep{gulwani2015inductive, muggleton2012ilp, kitzelmann2009inductive} mention about the systems and applications of ILP in interdisciplinary
areas. We next give related work of ILP in general
and then focus on ILP applied in the field of Natural Language Processing (NLP).

The foundations of ILP can be found in research by Plotkin \citep{plotkin1970note, plotkin1971further},
Shapiro \citep{shapiro1983algorithmic} and Sammut and Banerji \citep{sammut1986learning}.
The founding paper of Muggleton \citep{muggleton1991inductive} led to the launch of the first international workshop on ILP.
The strength of ILP lay in its ability to draw on and extend the existing successful paradigms of ML and
Logic Programming.
At the beginning, ILP was associated with the introduction of foundational theoretical concepts which included Inverse Resolution
\citep{muggleton1992machine, muggleton1995inverse} and Predicate Invention \citep{muggleton1992machine, muggleton1991inductive}.
A number of ILP systems were developed along with learning about the theoretical concepts of ILP such as FOIL \citep{quinlan1990learning} and
Golem \citep{muggleton1990efficient}.
The widely-used ILP system Progol \citep{muggleton1995inverse} introduced a new logically-based approach to refinement graph search of the hypothesis
space based on inverting the entailment relation. Meanwhile, the TILDE system \citep{de1997logical} demonstrated the efficiency which could be gained by
upgrading decision-tree learning algorithms to first-order logic, this was soon extended towards other ML problems.
Some limitations of Prolog-based ILP include requiring extensional background and negative examples,
lack of predicate invention, search limitations and inability to handle cuts.
Integrating bottom-up and top-down searches, incorporating predicate
invention, eliminating the need for explicit negative examples and allowing restricted use of cuts helps in solving these issues \citep{mooney1996inductive}.

Probabilistic ILP (PILP) also gained popularity \citep{muggleton1996stochastic, cussens2001integrating, de2008probabilistic}, its Prolog-based
systems such as PRISM \citep{sato2005generative} and FAM \citep{cussens2001parameter} separate the actual learning of the logic program from the
probabilistic parameters estimation of the individual clauses. However in practice, learning the structure and parameters of probabilistic logic
representation simultaneously has proven to be a challenge \citep{muggleton2002learning}.
PILP is mainly a unification of the probabilistic reasoning of Machine Learning with the relational logical representations offered by ILP.

Meta-interpretive learning (MIL) \citep{muggleton2014meta} is a recent ILP method which learns recursive definitions using Prolog and ASP-based declarative representations.
MIL is an extension of the Prolog meta-interpreter; it derives a proof by repeatedly fetching the first-order Prolog clauses and additionally fetching higher-order meta-rules
whose heads unify with a given goal, and saves the resulting meta-substitutions to form a program.

Most ILP research has been aimed at Horn programs which exclude Negation as Failure (NAF).
Negation is a key feature of logic programming and provides a
means for monotonic commonsense reasoning under incomplete information. This fails to exploit the full potential of normal programs that allow NAF.

We next give an overview of ILP systems based on ASP
that are designed to operate in the presence of negation.
Then we give an overview of ILP literature related to NLP.

\subsection{ASP-based ILP Systems}

The \emph{eXtended Hybrid Abductive Inductive Learning} system (XHAIL)
is an ILP approach based on ASP
that generalises techniques of language and search bias from Horn clauses to normal logic programs with full usage of NAF \citep{ray2009nonmonotonic}.
Like its predecessor system
Hybrid Abductive Inductive Learning (HAIL)
which operated on Horn clauses,
XHAIL is based on Abductive Logic Programming (ALP) \citep{Kakas1992},
we give more details on XHAIL in Section~\ref{secExtendingXHAIL}.

The \emph{Incremental Learning of Event Definitions} (ILED)
algorithm \citep{katzouris2015incremental}
relies on Abductive-Inductive learning and comprises of a scalable clause refinement methodology based on a compressive summarization of clause coverage in a stream of examples.
Previous ILP learners were batch learners and required all training data to be in place prior to the initiation of the learning process.
ILED learns incrementally by processing training instances when they become available and altering previous inferred knowledge to fit new observation, this is also known as theory revision.
It exploits previous computations to speed-up the learning since revising the hypothesis is considered more efficient than learning from scratch.
ILED attempts to cover a maximum of examples by re-iterating
over previously seen examples when the hypothesis has been
refined.
While XHAIL can ensure optimal example coverage easily
by processing all examples at once,
ILED does not preserve this property due to a non-global view on examples.

When considering ASP-based ILP,
negation in the body of rules is not the only interesting
addition to the overall concept of ILP.
An ASP program can have several independent solutions,
called answer sets, of the program.
Even the background knowledge $B$ can admit several
answer sets without any addition of facts from examples.
Therefore, a hypothesis $H$
can cover some examples in one answer set,
while others are covered by another answer set.
XHAIL and ILED approaches are based on finding a hypothesis
that is covering  all examples in a \emph{single} answer set.

The \emph{Inductive Learning of Answer Set Programs} approach
(ILASP) is an extension of the notion of learning from answer sets \citep{law2014inductive}.
Importantly, it covers positive examples bravely (i.e., in at least one answer set)
and ensures that the negation of negative examples is cautiously entailed (i.e., no negative example is covered in any answer set).
Negative examples are needed to learn Answer Set Programs with non-determinism otherwise there is no concept of what should \emph{not} be in an Answer Set.
ILASP conducts a search in multiple stages for brave and cautious entailment and processes all examples at once.
ILASP performs a less informed hypothesis search than XHAIL or ILED, that means large hypothesis spaces are infeasible for ILASP while they are not problematic for XHAIL and ILED,
on the other hand, ILASP supports aggregates and constraints
while the older systems do not support these.
ILASP2 \citep{law2015learning} extends the hypothesis
space of ILASP with choice rules and weak constraints.
This permits searching for hypotheses that encode
preference relations.

\subsection{ILP and NLP}

From NLP point of view, the hope of ILP is to be able to steer a mid-course between these two alternatives of large-scale but shallow levels
of analysis and small scale but deep and precise analysis. ILP should produce a better ratio between breadth of coverage and depth of analysis
\citep{muggleton1999inductive}.
ILP has been applied to the field of NLP successfully; it has not only been shown to have higher accuracies than various other
ML approaches in learning the past tense of English but also shown to be capable of learning accurate grammars
which translate sentences into deductive database queries \citep{law2014inductive}.

Except for one early application \citep{wirth1989completing} no application of ILP methods surfaced until the system
CHILL \citep{mooney1996inductive} was developed which learned a shift-reduce parser in Prolog from a training corpus of sentences paired with the desired
parses by learning control rules and uses ILP to learn control strategies within this framework. This work
also raised several issues regarding the capabilities and testing of ILP systems. CHILL was also used for parsing database queries to automate the
construction of a natural language interface \citep{zelle1996learning} and helped in demonstrating its ability to learn semantic mappings as well.

An extension of CHILL, CHILLIN \citep{zelle1994combining} was used along with an extension of FOIL, mFOIL \citep{tang2001using} for semantic parsing. Where
CHILLIN combines top-down and bottom-up induction methods and mFOIL is a top-down ILP algorithm designed keeping imperfect data in mind, which portrays
whether a clause refinement is significant for the overall performance with the help of
a pre-pruning algorithm. This emphasised on how the combination of
multiple clause constructors helps improve the overall learning; which is a rather similar concept to Ensemble Methods in standard ML.
Note that CHILLIN pruning is based on probability estimates and has the purpose of dealing with inconsistency in the data.
Opposed to that, XHAIL already supports learning from inconsistent data, and the pruning we discuss in Section~\ref{secPruning} aims to increase scalability.

Previous work ILP systems such as TILDE and Aleph \citep{srinivasan2001aleph} have been applied to preference learning which addressed learning ratings such as good, poor and bad. ASP
expresses preferences through weak constraints and may also contain weak constraints or optimisation statements which impose an ordering on the answer sets
\citep{law2015learning}.

The system of Mitra and Baral~\citep{mitra2016addressing} uses ASP as primary knowledge representation and reasoning language
to address the task of Question Answering.
They use a rule layer that is partially learned with XHAIL
to connect results from an Abstract Meaning Representation parser and an Event Calculus theory as background knowledge.

\section{Extending XHAIL algorithm and system}
\label{secExtendingXHAIL}

Initially, we intended to use the latest ILP systems
(ILASP2 or ILED) in our work.
However,
preliminary experiments with ILASP2
showed a lack in scalability (memory usage)
even for only 100 sentences due to the unguided hypothesis search space.
Moreover, experiments with ILED
uncovered several problematic corner cases in the ILED algorithm that led to empty hypotheses when processing examples
that were mutually inconsistent
(which cannot be avoided in real-life NLP data).
While trying to fix these problems in the algorithm,
further issues in the ILED implementation came up.
After consulting the authors of \citep{mitra2016addressing}
we learned that they had the same issues and used XHAIL,
therefore we also opted to base our research on XHAIL
due to it being the most robust tool for our task in comparison to the others.

\xfw{tbph}
Although XHAIL is applicable,
we discovered several drawbacks
and improved the approach and the XHAIL system.
We provide an overview of the parts we changed
and then present our modifications. Figure~\ref{figXhail} shows in the middle
the original XHAIL components and on the right
our extension.

XHAIL finds a hypothesis using several steps.
Initially the examples $E$ plus background knowledge $B$
are transformed into a theory of Abductive Logic Programming \citep{Kakas1992}.
The \emph{Abduction} part of XHAIL explains observations
with respect to a prior theory,
which yields the \emph{Kernel Set}, $\Delta$.
$\Delta$ is a set of potential heads of rules given by $M$
such that a maximum of examples $E$ is satisfied
together with $B$.
\begin{example}[continued]
Given $(M,E,B)$ from Example~\ref{exMini},
XHAIL uses $B$, $E$, and the head part of $M$,
to generate the Kernel Set $\Delta$ by abduction.
\begin{equation*}
\Delta = \left \{
  \begin{tabular}{lll}
  flies(a)\\
flies(b)\\
flies(c)
  \end{tabular}
\right \}
\end{equation*}
\end{example}

The \emph{Deduction} part
uses $\Delta$ and the body part of the mode bias $M$
to generate a ground program $K$.
$K$ contains rules which define atoms in $\Delta$ as true
based on $B$ and $E$.

The \emph{Generalisation} part replaces constant terms in $K$
with variables according to the mode bias $M$,
which yields a non-ground program $K'$.
\begin{example}[continued]
\label{exKKPrime}
From the above $\Delta$ and $M$ from \eqref{eqM},
deduction and generalisation yield the following $K$ and $K'$.
\begin{align*}
K &= \left \{
  \begin{tabular}{lll}
flies(a) :- bird(a),\, \naf penguin(a) \\
flies(b) :- bird(b),\, \naf penguin(b) \\
flies(c) :- bird(c),\, \naf penguin(c)
  \end{tabular}
\right \} \\
K' &= \left \{
  \begin{tabular}{lll}
flies(X) :- bird(X),\, \naf penguin(X)\\
flies(Y) :- bird(Y),\, \naf penguin(Y)\\
flies(Z) :- bird(Z),\, \naf penguin(Z)
  \end{tabular}
\right \}
\end{align*}
\end{example}

The \emph{Induction} part
searches for the smallest part of $K'$
that entails as many examples of $E$ as possible
given $B$.
This part of $K'$ which can contain a subset of the rules of $K'$
and for each rule a subset of body atoms
is called a hypothesis $H$.
\begin{example}[continued]
The smallest hypothesis that covers all examples $E$
in \eqref{eqE} is \eqref{eqExHypothesis}.
\end{example}

We next describe our modifications of XHAIL.

\subsection{Kernel Pruning according to Support}
\label{secPruning}

The computationally most expensive
part of the search in XHAIL is Induction.
Each non-ground rule in $K'$
is rewritten into a combination of several guesses,
one guess for the rule and one additional guess
for each body atom in the rule.

We moreover observed
that some non-ground rules in $K'$ are generalisations
of many different ground rules in $K$,
while some non-ground rules correspond with only
a single instance in $K$.
In the following,
we say that the \emph{support of $r$ in $K$}
is the number of ground rules in $K$
that are transformed into $r \ins K'$
in the Generalisation module of XHAIL
(see Figure~\ref{figXhail}).

Intuitively, the higher the support,
the more examples can be covered with that rule,
and the more likely that rule or a part of it
will be included in the optimal hypothesis.

Therefore we modified the XHAIL algorithm as follows.
\begin{itemize}
\item
  During Generalisation, we keep track of the support
  of each rule $r \ins K'$ by counting
  how often a generalisation yields the same rule $r$.
\item
  We add an integer pruning parameter $Pr$ to the algorithm
  and use only those rules from $K'$ in the
  Induction component that have a support higher
  than $Pr$.
\end{itemize}
This modification is depicted as bold components
which replace the dotted Generalisation module
in Figure~\ref{figXhail}.

Pruning has several consequences.
From a theoretical point of view,
the algorithm becomes incomplete for $Pr \gts 0$,
because Induction searches in a subset of the
relevant hypotheses.
Hence Induction might not be able to find
a hypothesis that covers all examples, although
such a hypothesis might exist with $Pr \eqs 0$.
From a practical point of view,
pruning realises something akin to regularisation
in classical ML;
only strong patterns in the data will find their way
into Induction and have the possibility
to be represented in the hypothesis.
A bit of pruning will therefore automatically
prevent overfitting and generate more general hypotheses.
As we will show in Experiments in Section~\ref{results},
the pruning allows to configure a trade-off
between considering low-support rules instead of omitting them entirely,
as well as,
finding a more optimal hypothesis in comparison to a highly
suboptimal one.

\subsection{Unsat-core based and Best-effort Optimisation}
\label{secUnsatBestEffort}

We observed that ASP search in XHAIL Abduction and Induction components
progresses very slowly from a suboptimal
to an optimal solution.
XHAIL integrates version 3 of Gringo~\citep{Gebser2011gringo3}
and Clasp~\citep{Gebser2012aij}
which are both quite outdated.
In particular Clasp in this version does not support
three important improvements
that have been found for ASP optimisation:
\begin{inparaenum}[(i)]
\item unsat-core optimisation~\citep{Andres2012},
\item stratification for obtaining suboptimal answer sets~\citep{Ansotegui2013maxsat,Alviano2015aspopt}, and
\item unsat-core shrinking~\citep{Alviano2016coreshrinking}.
\end{inparaenum}

Method (i) inverts the classical branch-and-bound search methodology which progresses from worst to better solutions.
Unsat-core optimisation assumes all costs can be avoided
and finds unsatisfiable cores of the problem
until the assumption is true and a feasible solution is found.
This has the disadvantage of providing only the final optimal solution,
and to circumvent this disadvantage, stratification in method (ii) was developed which allows for combining branch-and-bound with
method (i) to approach the optimal value both from cost 0 and from infinite cost.
Furthermore, unsat-core shrinking in method (iii),
also called `anytime ASP optimisation',
has the purpose of providing suboptimal solutions and
aims to find smaller cores which can speed up the search
significantly by cutting more of the search space (at the cost of searching for a smaller core).
In experiments with the inductive encoding of XHAIL
we found that all three methods have a beneficial effect.

Currently, only the WASP solver~\citep{DBLP:conf/lpnmr/AlvianoDFLR13,DBLP:conf/lpnmr/AlvianoDLR15}
supports all of (i), (ii), and (iii),
therefore we integrated WASP into XHAIL,
which has a different output than Clasp.
We also upgraded XHAIL to use Gringo version 4
which uses the new ASP-Core-2 standard
and has some further (performance) advantages over older versions.

Unsat-core optimisation often finds solutions
with a reasonable cost, near the optimal value,
and then takes a long time to find the true optimum
or prove optimality of the found solution.
Therefore, we extended XHAIL as follows:
\begin{itemize}
\item a time budget for search can be specified on the command line,
\item after the time budget is elapsed the best-known solution at that point is used and the algorithm continues,
furthermore
\item the distance from the optimal value is provided as output.
\end{itemize}
This affects the Induction step
in Figure~\ref{figXhail}
and introduces a best-effort strategy;
along with the obtained hypothesis
we also get the distance
from the optimal hypothesis,
which is zero for optimal solutions.

Using a suboptimal hypothesis means,
that either fewer examples
are covered by the hypothesis than possible,
or that the hypothesis is bigger than necessary.
In practice,
receiving a result is better than receiving no result at all,
and our experiments show that XHAIL becomes applicable
to reasonably-sized datasets using these extensions.

\subsection{Other Improvements}

We made two minor
engineering contributions to XHAIL.
A practically effective improvement of XHAIL
concerns $K'$.
As seen in Example~\ref{exKKPrime},
three rules that are equivalent modulo variable renaming
are contained in $K'$.
XHAIL contains canonicalization algorithms for avoiding
such situations, based on hashing body elements of rules.
However, we found that for cases with more than one variable
and for cases with more than one body atom,
these algorithms are not effective because XHAIL
\begin{inparaenum}[(i)]
\item uses a set data structure that maintains an order over elements,
\item the set data structure is sensitive to insertion order, and
\item hashing the set relies on the order to be canonical.
\end{inparaenum}
We made this canonicalization algorithm applicable
to a far wider range of cases
by changing the data type of rule bodies
in XHAIL to a set that maintains an order
depending on the value of set elements.
This comes at a very low additional cost
for set insertion and often reduces size of $K'$
(and therefore computational effort for Induction step)
without adversely changing the result of induction.

Another improvement concerns debugging the ASP solver.
XHAIL starts the external ASP solver and waits for the result.
During ASP solving, no output is visible,
however, ASP solvers provide output
that is important for tracking the distance from
optimality during a search.
We extended XHAIL so that the output of the ASP solver
can be made visible during the run
using a command line option.

\work{bth}
\section{Chunking with ILP} \label{chunk}

We evaluate the improvements of the previous section
using the NLP task of chunking.
Chunking \citep{tjong2000introduction} or shallow parsing is the
identification of short phrases such as noun phrases or prepositional phrases, usually based heavily on Part of Speech (POS) tags.
POS provides only information about the token type,
i.e., whether words are nouns, verbs, adjectives, etc.,
and chunking derives from that a shallow phrase structure,
in our case a single level of chunks.

Our framework for chunking has three main parts
as shown in Figure~\ref{table:Work}.
Preprocessing is done using the Stanford CoreNLP tool from which we obtain the facts
that are added to the background knowledge of XHAIL
or used with a hypothesis to predict the chunks of an input.
Using XHAIL as our ILP solver
we learn a hypothesis (an ASP program) from the background knowledge, mode bias, and from examples
which are generated using the gold-standard data.
We predict chunks using our learned hypothesis and facts from preprocessing, using the Clingo \citep{gebser2008user} ASP solver.
We test by scoring predictions against gold chunk annotations.

\begin{example}
An example sentence in the SemEval iSTS dataset \citep{companionpaper} is as follows.
\begin{align}
\label{eqExampleSentence}
\textit{Former Nazi death camp guard Demjanjuk dead at 91}
\end{align}
The chunking present in the SemEval gold standard is as follows.
\smallskip
\begin{align}
\label{eqExampleGoldchunks}
\textit{[ Former Nazi death camp guard Demjanjuk ] [ dead ] [ at 91 ]}
\end{align}
\end{example}

\subsection{Preprocessing}
Stanford CoreNLP tools \citep{manning2014stanford} are used for tokenisations and POS-tagging of the input. Using a shallow parser
\citep{bohnet2013joint} we obtain the dependency relations for the sentences.
\todo[inline,disable]{P:why CONLL? -- M: Mentioned why we use CONLL format, it even has its own chunking competitions in the past (http://www.aclweb.org/anthology/W00-0726) -- P: but it is irrelevant here, because we could just convert output of Stanford directly to ASP, so I removed this reference}
Our ASP representation contains atoms of the following form:
\begin{itemize}
\item
  $\mi{pos(P,T)}$ which represents that token $T$ has POS tag $P$,
\item
  $\mi{form(T,Text)}$ which represents that token $T$ has surface form $\mi{Text}$,
\item
  $\mi{head(T_1,T_2)}$ and $\mi{rel(R,T)}$
  which represent that token $T_2$ depends on token $T_1$
  with dependency relation $R$.
\end{itemize}

\begin{figure}
\centering
\begin{subfigure}{1\linewidth}
  \centering
  \begin{lstlisting}[
    basicstyle=\small, %
    %
    language=Prolog,
    frame=single
  ]
  pos(c_NNP,1). head(2,1). form(1,"Former"). rel(c_NAME,1).
  pos(c_NNP,2). head(5,2). form(2,"Nazi"). rel(c_NMOD,2).
  pos(c_NN,3). head(4,3). form(3,"death"). rel(c_NMOD,3).
  pos(c_NN,4). head(5,4). form(4,"camp"). rel(c_NMOD,4).
  pos(c_NN,5). head(7,5). form(5,"guard"). rel(c_SBJ,5).
  pos(c_NNP,6). head(5,6). form(6,"Demjanjuk"). rel(c_APPO,6).
  pos(c_VBD,7). head(root,7). form(7,"dead"). rel(c_ROOT,7).
  pos(c_IN,8). head(7,8). form(8,"at"). rel(c_ADV,8).
  pos(c_CD,9). head(8,9). form(9,"91"). rel(c_PMOD,9).
  \end{lstlisting}
  \vspace*{-0.8em}
  \caption{Preprocessing Output}
  \label{figPrepro}
  \vspace{0.25cm}
\end{subfigure}
\begin{subfigure}{1\linewidth}
  \centering
  \begin{lstlisting}[
    basicstyle=\small, %
    %
    language=Prolog,
    frame=single
  ]
  postype(X) :- pos(X,_).
  token(X) :- pos(_,X).
  nextpos(P,X) :- pos(P,X+1).
  \end{lstlisting}
  \vspace*{-0.8em}
  \caption{Background Knowledge}
  \label{figBackground}
  \vspace{0.25cm}
\end{subfigure}
\begin{subfigure}{1\linewidth}
  \centering
  \begin{lstlisting}[
    basicstyle=\small, %
    %
    language=Prolog,
    frame=single
  ]
  #modeh split(+token).
  #modeb pos($postype,+token).
  #modeb nextpos($postype,+token).
  \end{lstlisting}
  \vspace*{-0.8em}
  \caption{Mode Restrictions}
  \label{figModes}
  \vspace{0.25cm}
\end{subfigure}
\begin{subfigure}{1\linewidth}
  \centering
  \begin{lstlisting}[
    basicstyle=\small, %
    %
    language=Prolog,
    frame=single
  ]
  goodchunk(1) :- not split(1), not split(2), not split(3),
                  not split(4), not split(5), split(6).
  goodchunk(7) :- split(6), split(7).
  goodchunk(8) :- split(7), not split(8).
  #example goodchunk(1).
  #example goodchunk(7).
  #example goodchunk(8).
  \end{lstlisting}
  \vspace*{-0.8em}
  \caption{Examples}
  \label{figExamples}
\end{subfigure}
\caption{XHAIL input for the sentence 'Former Nazi death camp guard Demjanjuk dead at 91' from the Headlines Dataset}
\label{figure:input}
\end{figure}

\begin{example}[continued]
Figure~\ref{figPrepro} shows the result of
preprocessing performed on
sentence~\eqref{eqExampleSentence},
which is a set of ASP facts.
\end{example}
We use Penn Treebank POS-tags as they are provided by Stanford CoreNLP.
To form valid ASP constant terms from POS-tags,
we prefix them with `c\_', replace special characters with lowercase letters
(e.g., `PRP\$' becomes `c\_PRPd').
In addition, we create specific POS-tags for punctuation (see Section~\ref{secResults}).

\subsection{Background Knowledge and Mode Bias}
\label{secBM}
Background Knowledge we use is shown in
Figure~\ref{figBackground}.
We define which POS-tags can exist in predicate $\mi{postype}/1$
and which tokens exist in predicate $\mi{token}/1$.
Moreover, we provide for each token
the POS-tag of its successors token in predicate $\mi{nextpos}/2$.

Mode bias conditions are shown in Figure~\ref{figModes},
these limit the search space for hypothesis generation.
Hypothesis rules contain as head atoms of the form
\begin{align*}
  \mi{split}(T)
\end{align*}
which indicates,
that a chunk ends at token $T$ and a new chunk
starts at token $T\,{+}\,1$.
The argument of predicates $\mi{split}/1$ in the head
is of type $\mi{token}$.

The body of hypothesis rules can contain
$\mi{pos}/2$ and $\mi{nextpos}/2$ predicates,
where the first argument is a constant of type $\mi{postype}$
(which is defined in Figure~\ref{figBackground})
and the second argument is a variable of type $\mi{token}$.
Hence this mode bias searches for rules defining chunk splits
based on POS-tag of the token and the next token.

We deliberately use a very simple mode bias
that does not make use of all atoms in the facts
obtained from preprocessing.
This is discussed in Section~\ref{secDiscussion}.

\subsection{Learning with ILP}
Learning with ILP is based on examples that guide the search.
Figure~\ref{figExamples} shows rules that recognise
gold standard chunks and $\mi{\#example}$ instructions
that define for XHAIL
which atoms must be true to entail an example.
These rules with $\mi{goodchunk}/1$ in the head
define what a good (i.e., gold standard) chunk is
in each example based on where a split in a chunk occurs
in the training data
to help in the learning of a hypothesis for chunking.

Note that negation is present only in these rules,
although we could use it anywhere else
in the background knowledge.
Using the background knowledge, mode bias, and examples,
XHAIL is then able to learn a hypothesis.

\subsection{Chunking with ASP using Learned Hypothesis}
The hypothesis generated by XHAIL
can then be used together with the background knowledge
specified in Figure~\ref{figBackground},
and with the preprocessed input of a new sentence.
Evaluating all these rules yields a set of split points
in the sentence, which corresponds
to a predicted chunking of the input sentence.
\begin{example}[continued]
  Given sentence~\eqref{eqExampleSentence}
  with token indices $1, \ldots, 9$,
  an answer set that contains the
  atoms $\{ \mi{split}(6), \mi{split}(7) \}$
  and no other atoms for predicate $\mi{split}/1$
  yields the chunking shown in~\eqref{eqExampleGoldchunks}.
\end{example}

\section{Evaluation and Discussion} \label{results}

\subsection{Datasets}
We are using the datasets from the SemEval 2016 iSTS Task 2 \citep{companionpaper}, which included two separate files containing sentence pairs.
Three different datasets were provided: Headlines, Images, and Answers-Students. The Headlines dataset was mined by various news sources by
European Media Monitor.
The Images dataset was a collection of captions obtained from the Flickr dataset \citep{rashtchian2010collecting}.
The Answers-Students corpus consists of the interactions between students and the BEETLE II tutorial dialogue system which is an intelligent tutoring
engine that teaches students in basic electricity and electronics.
In the following, we denote S1 and S2,
by sentence 1 and sentence 2 respectively,
of sentence pairs in these datasets.
Regarding the size of the SemEval Training dataset,
Headlines and Images datasets are larger and contained 756 and 750 sentence pairs, respectively. However,
the Answers-Students dataset was smaller and contained only 330 sentence pairs.
In addition, all datasets contain a Test portion
of sentence pairs.

\revA{%
We use $k$-fold cross-validation to evaluate chunking with ILP,
which yields $k$ learned hypotheses and $k$ evaluation scores for each parameter setting.
We test each of these hypotheses also on the Test portion of the respective dataset.
From the scores obtained this way
we compute mean and standard deviation,
and perform statistical tests to find out whether
observed score differences between parameter settings
is statistically significant.

Table~\ref{table:Split} shows which portions of the
SemEval Training dataset we used for 11-fold cross-validation.
In the following, we call these datasets
\emph{Cross-Validation Sets}.
We chose the first 110 and 550 examples to use for 11-fold
cross-validation which results in training set sizes 100 and 500, respectively.
As the Answers-Students dataset was smaller,
we merged its sentence pairs in order to obtain
a Cross-Validation Set size of 110 sentences,
using the first 55 sentences from S1 and S2;
and for 550 sentences,
using the first 275 sentences from S1 and S2 each.}
As Test portions we only use the original SemEval Test datasets and we always test S1 and S2 separately.

\DataSplit{tbh}

\subsection{Scoring}

We use difflib.SequenceMatcher in Python
to match the sentence chunks obtained from learning in ILP
against the gold-standard sentence chunks.
From the matchings obtained this way, we compute precision, recall, and F1-score as follows.
\begin{align*}
 Precision &= \dfrac{\textnormal{No. of Matched Sequences}}{\textnormal{No. of ILP-learned Chunks}} \\[1ex]
 Recall &= \dfrac{\textnormal{No. of Matched Sequences}}{\textnormal{No. of Gold Chunks}} \\[1ex]
 Score &= 2 \times \dfrac{\textnormal{Precision} \times \textnormal{Recall}}{\textnormal{Precision} + \textnormal{Recall}}
\end{align*}

To investigate the effectivity of our mode bias
for learning a hypothesis that can correctly classify
the dataset,
\revA{we perform cross-validation (see above)
and measure correctness of all hypotheses obtained in
cross-validation also on the Test set.}

Because of differences in S1/S2 portions of datasets,
we report results separately for S1 and S2.
We also evaluate classification separately for S1 and S2
for the Answers-Students dataset,
although we train on a combination of S1 and S2.

\subsection{Experimental Methodology}
\label{secExperimentalMethodology}
We use Gringo version~4.5 \citep{Gebser2011gringo3}
and we use WASP version~2 (Git hash a44a95)
\citep{DBLP:conf/lpnmr/AlvianoDLR15}
configured to use unsat-core optimisation
with disjunctive core partitioning,
core trimming,
a budget of 30 seconds for computing the first answer set
and for shrinking unsatisfiable cores with progressive
shrinking strategy.
These parameters were found most effective
in preliminary experiments.
We configure our modified XHAIL solver to allocate
a budget of 1800~seconds for the Induction part
which optimises the hypothesis
(see Section~\ref{secUnsatBestEffort}).
Memory usage never exceeded 5~GB.

Tables~\ref{table:Headlines}--\ref{table:StudentAns}
contains the experimental results for each Dataset,
where columns
Size, $Pr$, and $So$ respectively,
show the \revA{number of sentences} used to learn the hypothesis,
the pruning parameter for generalising the learned hypothesis (see Section~\ref{secPruning}),
and the rate of how close the learned hypothesis is to the optimal result, respectively.
$So$ is computed according to the following formula: $\mi{\mathit{So}}\eqs\frac{Upper bound - Lower bound}{Lower bound}$
, which is based on upper and lower bounds on the cost of the answer set. An $So$ value of zero means optimality,
and values above zero mean suboptimality;
so the higher the value, the further away from optimality.
Our results comprise of \revA{the mean and standard deviation of the F1-scores obtained from our 11-fold cross-validation test set of S1 and S2 individually
(column CV).} Due to lack of space, we opted to leave out the scores of precision and recall, but these values show similar trends as in the Test set.
For the Test sets of both S1 and S2, we \revA{include the mean and standard deviation} of the Precision, Recall and F1-scores (column group T).

\revA{%
When testing machine-learning based systems,
comparing results obtained on a single test set is often
not sufficient,
therefore we performed cross-validation
to obtain mean and standard deviation about our benchmark metrics.
To obtain even more solid evidence
about the significance of the measured results,
we additionally performed a one-tailed paired t-test
to check if a measured F1 score
is significantly higher in one setting than in another one.
We consider a result significant if $p < 0.05$, i.e.,
if there is a probability of less than 5~\% that the result is due to chance.
Our test is one-tailed because we check whether one result
is higher than another one, and it is a paired test
because we test different parameters on the same set of
11 training/test splits in cross-validation.
There are even more powerful methods
for proving significance of results
such as bootstrap sampling~\citep{Efron1986},
however these methods require markedly higher
computational effort in experiments
and our experiments already show significance
with the t-test.}

Rows of Tables~\ref{table:Headlines}--\ref{table:StudentAns}
contain results for learning from
\revA{100 resp.\ 500 example sentences},
and for different pruning parameters.
For both learning set sizes,
we increased pruning stepwise starting from value 0
until we found an optimal hypothesis ($\mi{\mathit{So}}\eqs0$)
or until we saw a clear peak in classification score
\revA{in cross-validation}
(in that case, increasing the pruning is pointless
because it would increase optimality of the hypothesis
but decrease the prediction scores).

Note that datasets have been tokenised very differently,
and that also state-of-the-art systems in SemEval
used separate preprocessing methods for each dataset.
We follow this strategy to allow a fair comparison.
One example for such a difference is the Images dataset,
where the \quo{.} is considered as a separate token
and is later defined as a separate chunk,
however in Answers-Students dataset
it is integrated onto neighboring tokens.

\subsection{Results} \label{secResults}

We first discuss the results
of experiments with varying training set size
and varying pruning parameter,
then compare our approach with the state-of-the-art
systems,
and finally inspect the optimal hypotheses.

\paragraph{\revA{Training Set Size and Pruning Parameter}}
Tables~\ref{table:Headlines}--\ref{table:StudentAns}
show results of experiments,
where T denotes the Test portion of the respective dataset.

We observe that by increasing the size of the training set to learn the hypothesis, our scores improved considerably.
Due to more information being provided, the learned hypothesis can predict with higher F1 score.
We also observed that for the smaller training set size (100 sentences), lower pruning numbers (in rare cases even $Pr{=}0$) resulted in achieving
the optimal solution.
For a bigger training set size (500 sentences),
\revA{%
without pruning the ILP procedure does not find solutions close to the optimal solution.
However, by using pruning values up to $Pr{=}10$
we can reduce the size of the search space and find hypotheses
closer to the optimum,
which predict chunks with a higher F1 score.
Our statistical test shows that, in many cases,
several increments of the $Pr$ parameter yield significantly
better results, up to a point where prediction accuracy
degrades because too many examples are pruned away.}
To select the best hypothesis, we increase the pruning
parameter $\mi{Pr}$ until we reach the peak in the
F1 score in cross-validation.

Finding optimal hypotheses in the Inductive search
of XHAIL (where $\mathit{So}{=}0$)
is easily attained when learning from
100 sentences.
For learning from 500 sentences,
very higher pruning results in a trivial optimal
hypothesis (i.e., every token is a chunk)
which has no predictive power,
hence we do not increase $Pr$ beyond a value of 10.

Note that we never encountered timeouts
in the Abduction component of XHAIL,
only in the Induction part.
The original XHAIL tool without our improvements
yields only timeouts for learning from 500 examples,
and few hypotheses for learning from 100 examples.
Therefore we do not show these results in tables.

\FinalResults{tbh}
\paragraph{State-of-the-art comparison}
Table~\ref{table:final} shows a comparison of our results
with the baseline and the three best systems from the chunking subtask of Task 2 from SemEval2016 Task2 \citep{companionpaper}:
DTSim \citep{banjade2016dtsim}, FBK-HLT-NLP \citep{magnolini2016fbk} and runs 1 and 2 of IISCNLP
\citep{tekumalla2016iiscnlp}.
We also compare with results of our own system \quo{Inspire-Manual}~\citep{kazmi2016inspire}.
\revA{%
\begin{itemize}
 \item The baseline makes use of the automatic probabilistic chunker from the IXA-pipeline which provides Perceptron models~\citep{collins2002discriminative}
 for chunking and is trained on CONLL2000 corpora and corrected manually,
 \item DTSim uses a Conditional Random Field (CRF) based chunking tool using only POS-tags as features,
 \item FBK-HLT-NLP obtains chunks using a Python implementation of MBSP chunker which uses a Memory-based part-of-speech tagger generator~\citep{daelemans1996mbt},
 \item Run 1 of IISCNLP uses OpenNLP chunker which divides the sentence into syntactically correlated parts of words, but does not specify their internal
 structure, nor their role in the main sentence. Run 2 uses Stanford NLP Parser to create parse trees and then uses a perl script to create chunks based
 on the parse trees, and
 \item Inspire-Manual (our previous system) makes use of manually set chunking rules~\citep{abney1991parsing} using ASP \citep{kazmi2016inspire}.
\end{itemize}}

Using the gold-standard chunks provided by the organisers we were able to compute the precision, recall, and F1-scores for analysis on the
Headlines, Images and Answers-Students datasets.

For the scores of our system \quo{Inspire-Learned},
we used \revA{the mean and average of the best configuration of our system as obtained in cross-validation experiments on the Test set}
and compared against the other systems' Test set results.
Our system's performance is quite robust: it is always scores within the top three best systems.

\rules{tbh}
\paragraph{Inspection of Hypotheses}
Table~\ref{table:Rules} shows the rules that are obtained
from the hypothesis generated by XHAIL from Sentence 1 files of all the datasets. We have also tabulated the common rules present between the datasets and
the extra rules which differentiate the datasets from each other.
POS-tags for punctuation are `c\_p' for sentence-final
punctuation (`.', `?', and `!')
and `c\_c' for sentence-separating punctuation
(`,', `;', and `:').

Rules which occur in all learned hypotheses
can be interpreted as follows
(recall the meaning of $\mi{split}(X)$
from Section~\ref{secBM}):
\begin{inparaenum}[(i)]
\item chunks end at past tense verbs (VBD, e.g., `walked'),
\item chunks begin at subordinating conjunctions and prepositions (IN, e.g., `in'), and
\item chunks begin at 3rd person singular present tense verbs (VBZ, e.g., `walks').
\end{inparaenum}
Rules that are common to H and AS datasets are as follows:
\begin{inparaenum}[(i)]
\item chunks end at base forms of verbs (VB, e.g., `[to] walk'),
\item chunks begin at `to' prepositions (TO), and
\item chunks begin at past tense verbs (VBD).
\end{inparaenum}
The absence of (i) in hypotheses for the Images dataset
can be explained by the rareness of such verbs in
captions of images.
Note that (iii) together with the common rule (i)
means that all VBD verbs become separate chunks
in H and AS datasets.
Rules that are common to I and AS datasets are as follows:
\begin{inparaenum}[(i)]
\item chunks begin at non-3rd person verbs in present tense (VBP, e.g., `[we] walk'),
\item chunk boundaries are between a determiner (DT, e.g., `both') and a 3rd person singular present tense verb (VBZ), and
\item chunk boundaries are between adverbs (RB, e.g., `usually') and common, singular, or mass nouns (NN, e.g., `humor').
\end{inparaenum}
Interestingly, there are no rules common to H and I datasets except for the three rules mutual to all three datasets.

For rules occurring only in single datasets,
we only discuss a few interesting cases in the following.
Rules that are unique to the Headlines dataset
include rules which indicate that the sentence separators
`,', `;', and `:', become single chunks,
moreover chunks start at genitive markers (POS, `'s').
Both is not the case for the other two data sets.
Rules unique to the Images dataset
include that sentence-final punctuation (`.', `?', and `!')
become separate chunks,
rules for chunk boundaries
between verb (VB\_) and noun (NN\_) tokens,
and chunk boundaries between possessive pronouns
(PRP\$, encoded as `c\_PRPd', e.g., `their')
and participles/gerunds (VBG, e.g., `falling').
Rules unique to Answers-Students dataset
include chunks containing `existential there' (EX),
adverb tokens (RB),
gerunds (VBG),
and several rules for splits related to WH-determiners
(WDT, e.g., `which'),
WH-adverbs (WRB, e.g., `how'),
and prepositions (IN).

We see that learned hypotheses are interpretable,
which is not the case in
classical machine learning techniques such as
Neural Networks (NN), Conditional Random Fields (CRF),
and Support Vector Machines (SVM).

\subsection{\revA{Discussion}} \label{secDiscussion}
\revA{%
We next discuss the potential impact of our approach
in NLP and in other applications,
outline the strengths and weaknesses,
and discuss reasons for several design choices we made.}

\revA{%
\paragraph{Impact and Applicability}
ILP is applicable to many problems of traditional machine learning,
but usually only applicable for small datasets.
Our addition of pruning enables learning from larger datasets
at the cost of obtaining a more coarse-grained hypothesis
and potentially suboptimal solutions.

The main advantage of ILP is interpretability and that it can
achieve good results already with small datasets.
Interpretability of the learned rule-based hypothesis
makes the learned hypothesis transparent as opposed to
black-box models of other approaches in the field
such as Conditional Random Fields, Neural Networks, or
Support Vector Machines.
These approaches are often purely statistical, operate on
big matrices of real numbers instead of logical rules,
and are not interpretable.
The disadvantage of ILP is that it often does not achieve
the predictive performance of purely statistical approaches
because the complexity of ILP learning limits the number of
distinct features that can be used simultaneously.

Our approach allows finding suboptimal hypotheses
which yield a higher prediction accuracy
than an optimal hypothesis
trained on a smaller training set.
Learning a better model from a larger dataset is exactly
what we would expect in machine learning.
Before our improvement of XHAIL,
obtaining any hypothesis from larger datasets
was impossible:
the original XHAIL tool does not return any hypothesis
within several hours when learning from 500 examples.

Our chunking approach learns from a small portion
of the full SemEval Training dataset, based on only POS-tags,
but it still achieves results close to the state-of-the-art.
Additionally it provides an interpretable model
that allowed us to pinpoint non-uniform annotation practices
in the three datasets of the SemEval 2016 iSTS competition.}
These observations give direct evidence for
differences in annotation practice for three datasets
with respect to punctuation and genitives,
as well as differences in the content of the datasets

\revA{%
\paragraph{Strengths and weaknesses}
Our additions of pruning and the usage of suboptimal answer sets
make ILP more robust because it permits learning
from larger datasets
and obtaining (potentially suboptimal) solutions faster.

Our addition of a time budget
and usage of %
suboptimal answer sets
is a purely beneficial addition to the XHAIL approach.
If we disregard the additional benefits of pruning,
i.e., if we disable pruning by setting $Pr{=}0$,
then within the same time budget,
the same optimal solutions are to be found
as if using the original XHAIL approach.
In addition, before finding the optimal solution,
suboptimal hypotheses are provided in an online manner,
together with information about their
distance from the optimal solution.

The strength of pruning before the Induction phase
is, that it permits learning from a bigger set of examples,
while still considering all examples in the dataset.
A weakness of pruning is, that a hypothesis which fits
perfectly to the data might not be found anymore,
even if the mode bias could permit such a perfect fit.
In NLP applications this is not a big disadvantage,
because noise usually prevents a perfect fit anyways,
and overfitting models is indeed often a problem.
However, in other application domains such as
learning to interpret input data from user examples
\citep{gulwani2015inductive},
a perfect fit to the input data might be desired and required.
Note that pruning \emph{examples} to learn from inconsistent data
as done by Tang and Mooney \citep{tang2001using} is not necessary for our approach.
Instead, non-covered examples incur a cost that is optimised
to be as small as possible.%
}

\paragraph{\revA{Design decisions}}
In our study, we use a \emph{simple mode bias}
containing only the current and next POS tags,
which is a deliberate choice to make results
easier to compare.
We performed experiments with additional body atoms
$\mi{head/2}$ and $\mi{rel/2}$ in the body mode bias,
moreover with negation in the body mode bias.
However, these experiments yielded significantly
larger hypotheses with only small increases in accuracy.
Therefore we here limit the analysis to the simple case
and consider more complex mode biases as future work.
Note that the best state-of-the-art system (DTSim)
is a CRF model solely based on POS-tags,
just as our hypothesis is only making use of POS-tags.
By considering more than the current
and immediately succeeding
POS tag, DTSim can achieve better results than we do.

The \emph{representation of examples} is an important part
of our chunking case as described in Section~\ref{chunk}.
We define predicate $\mi{goodchunk}$
with rules that consider \emph{presence} and \emph{absence} of splits for each chunk.
We make use of the power of NAF in these rules.
We also experimented with an example representation
that just gave all desired splits as {\tt \#example split(X)} and all undesired splits as {\tt \#example not split(Y)}.
This representation contains an imbalance in the
split versus \naf split class,
moreover, chunks are not represented as a concept
that can be optimised in the inductive search
for the best hypothesis.
Hence, it is not surprising that this simpler representation
of examples gave drastically worse scores,
and we do not report any of these results in detail.

\section{Conclusion and Future Work} \label{conc}
\revA{%
Inductive Logic Programming combines logic programming
and machine learning, and it provides interpretable models, i.e., logical hypotheses, which are learned from data.
ILP has been applied to a variety of NLP and other problems
such as parsing \citep{zelle1996learning,tang2001using},
automatic construction of biological knowledge bases
from scientific abstracts \citep{Craven1999biokbc},
automatic scientific discovery \citep{King2004robotscientist},
and in Microsoft Excel \cite{gulwani2015inductive}
where users can specify data extraction rules using examples.
Therefore, ILP research has the potential for being used
in a wide range of applications.}

In this work, we explored the usage of ILP for the NLP task of chunking
and extend the XHAIL ILP solver to increase its scalability
and applicability for this task.
Results indicate that ILP is competitive to
state-of-the-art ML techniques for this task
and that we successfully extended XHAIL
to allow learning from larger datasets than previously
possible.
Learning a hypothesis using ILP has the advantage of
an interpretable representation of the learned knowledge,
such that we know exactly which rule has been learned by the program and how it affects our NLP task.
In this study, we also gain insights about the differences
and common points of datasets that we learned a hypothesis from.
Moreover, ILP permits learning from small training sets
where techniques such as Neural Networks fail to provide good results.

As a first contribution to the ILP tool XHAIL
we have upgraded the software so that it uses
the newest solver technology,
and that this technology is used in a best-effort manner
that can utilise suboptimal search results.
This is effective in practice,
because finding the optimal solution
can be disproportionately more difficult
than finding a solution close to the optimum.
Moreover, the ASP technique we use here provides
a clear information about the degree of suboptimality.
During our experiments, a new version of Clingo
was published which contains most techniques in WASP
(except for core shrinking).
We decided to continue using WASP for this study
because we saw that core shrinking is also beneficial to search.
Extending XHAIL to use Clingo
in a best-effort manner is quite straight-forward.

As a second contribution to XHAIL
we have added a \emph{pruning} parameter to the algorithm
that allows fine-tuning the search space for hypotheses
by filtering out rule candidates that are supported
by fewer examples than other rules.
This addition is a novel contribution to the algorithm,
which leads to significant improvements in efficiency,
and increases the number of hypotheses
that are found in a given time budget.
While pruning makes the method incomplete, it does not reduce expressivity. The hypotheses and background knowledge may still contain unrestricted
Negation as Failure. Pruning in our work is similar to the concept of the regularisation in ML and is there to prevent overfitting in the hypothesis generation.
\revA{%
Pruning enables the learning of
logical hypotheses with dataset sizes
that were not feasible before.
We experimentally observed a trade-off
between finding an optimal hypothesis that considers all potential rules on one hand,
and finding a suboptimal hypothesis that is based on rules that are supported by few examples.
Therefore the pruning parameter has to be adjusted
on an application-by-application basis.}

Our work has focused on providing comparable results to ML techniques and we have not utilised the full power of ILP with NAF in rule bodies and predicate invention.
As future work, we plan to extend the predicates usable
in hypotheses to provide a more detailed representation of the NLP task,
moreover we plan to enrich the background knowledge
to aid ILP in learning a better hypothesis with a deeper structure representing the boundaries
of chunks.

We provide the modified XHAIL system in a public repository fork \citep{xhailfork}.

\section*{Acknowledgments}

This research has been supported by
the Scientific and Technological Research Council of Turkey
(TUBITAK) [grant number 114E777] and by the Higher Education Commission
of Pakistan (HEC).
We are grateful to Carmine Dodaro for providing
us with support regarding the WASP solver.

\ifinlineref
\input{references.sty}
\else
\bibliography{bibliography}
\fi

\HeadlinesResults{ptb}
\ImagesResults{ptb}
\StudentAnsResults{ptb}

\end{document}